\documentclass{article}

\usepackage{microtype}
\usepackage{graphicx}
\usepackage{subcaption}
\usepackage{booktabs} 
\usepackage{tabularx}
\usepackage{ragged2e}
\usepackage{hyperref}

\usepackage[preprint]{icml2026}

\usepackage{amsmath}
\usepackage{amssymb}
\usepackage{mathtools}
\usepackage{amsthm}
\usepackage{multirow}

\usepackage[capitalize,noabbrev]{cleveref}
\usepackage{booktabs}
\usepackage{tabularx}
\usepackage{multirow}
\usepackage{makecell}
\usepackage{pifont}      
\usepackage[table]{xcolor}
\usepackage{colortbl}
\usepackage{threeparttable}
\usepackage{ragged2e}

\theoremstyle{plain}

\theoremstyle{definition}

\theoremstyle{remark}

\usepackage[textsize=tiny]{todonotes}

\icmltitlerunning{DeMUSE: Deep Multimodal Unified Sparse Experts for Dexterous Manipulation}

\hypersetup{
  pdftitle={Towards Dexterous Embodied Manipulation via Deep Multi-Sensory Fusion and Sparse Expert Scaling},
  pdfauthor={Yirui Sun; Guangyu Zhuge; Keliang Liu; Jie Gu; Zhihao Xia; Qionglin Ren; Chunxu Tian; Zhongxue Gan},
  pdfsubject={},  
  pdfkeywords={Embodied AI, Dexterous Manipulation, Multimodal Learning, Sensor Fusion, Sparse Experts}
}

\begin{document}

\twocolumn[
  \icmltitle{Towards Dexterous Embodied Manipulation \\via Deep Multi-Sensory Fusion and Sparse Expert Scaling}



  \icmlsetsymbol{equal}{*}

  \begin{icmlauthorlist}
    \icmlauthor{Yirui Sun}{fdu}
    \icmlauthor{Guangyu Zhuge}{hit}
    \icmlauthor{Keliang Liu}{fdu}
    \icmlauthor{Jie Gu}{fdu}
    \icmlauthor{Zhihao xia}{fdu}
    \icmlauthor{Qionglin Ren}{fdu}
    \icmlauthor{Chunxu tian}{fdu}
    \icmlauthor{Zhongxue Gan}{fdu}
  \end{icmlauthorlist}

  \icmlaffiliation{fdu}{College of Intelligent Robotics and Advanced Manufacturing, Fudan University}
  \icmlaffiliation{hit}{School of Information Science and Engineering, Harbin Institute of Technology}

  \icmlcorrespondingauthor{Chunxu tian}{chxtian@fudan.edu.cn}
  \icmlcorrespondingauthor{Zhongxue Gan}{ganzhongxue@fudan.edu.cn}

  \icmlkeywords{Embodied AI}

  \vskip 0.3in
]



\printAffiliationsAndNotice{}  

\begin{abstract}
  Realizing dexterous embodied manipulation necessitates the deep integration of heterogeneous multimodal sensory inputs. However, current vision-centric paradigms often overlook the critical force and geometric feedback essential for complex tasks. This paper presents DeMUSE, a Deep Multimodal Unified Sparse Experts framework leveraging a Diffusion Transformer to integrate RGB, depth, and 6-axis force into a unified serialized stream. Adaptive Modality-specific Normalization (AdaMN) is employed to recalibrate modality-aware features, mitigating representation imbalance and harmonizing the heterogeneous distributions of multi-sensory signals. To facilitate efficient scaling, the architecture utilizes a Sparse Mixture-of-Experts (MoE) with shared experts, increasing model capacity for physical priors while maintaining the low inference latency required for real-time control. A Joint denoising objective synchronously synthesizes environmental evolution and action sequences to ensure physical consistency. Achieving success rates of 83.2\% and 72.5\% in simulation and real-world trials, DeMUSE demonstrates state-of-the-art performance, validating the necessity of deep multi-sensory integration for complex physical interactions.
\end{abstract}

\begin{figure}[!t]
  \begin{center}
    \centerline{\includegraphics[width=0.9\columnwidth]{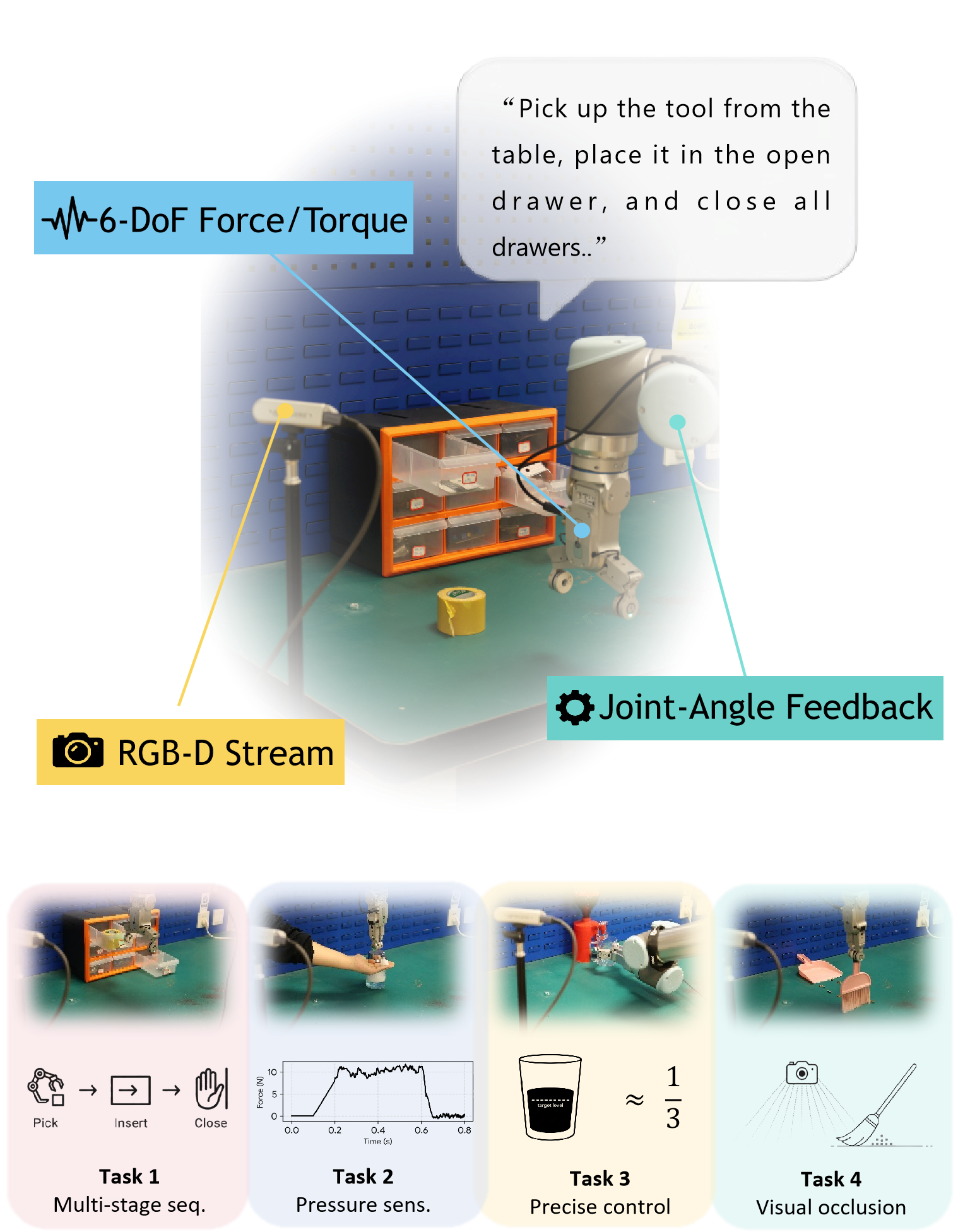}}
    \caption{
       The top image shows the layout of the experimental platform and the sources of multimodal inputs. The bottom image shows the tasks involved in real robot operation.
    }
    \vspace{-0.4in}
    \label{real}
  \end{center}
\end{figure}

\section{Introduction} \label{sec:intro}

Building general-purpose embodied agents represents a grand challenge in AI, requiring a seamless synergy between high-level semantic reasoning and low-level physical dexterity\cite{intelligence2504pi0,lin2025storm,xu2024survey}. Recent advancements in Vision-Language-Action (VLA)\cite{ghosh2024octo,openvla,intelligence2504pi0} models have achieved remarkable milestones in semantic generalization through large-scale pre-training, while Diffusion-based policies\cite{kim2026cosmos,liurdt,vpp} have demonstrated exceptional expressive power in capturing complex, multimodal action distributions. Despite these successes, current foundation models primarily focus on visual-semantic mappings, often treating the agent's interaction with the physical world as a purely visual task. However, to achieve truly precise and physically-consistent manipulation, there is an urgent need to move beyond vision-centric inputs and integrate a broader spectrum of sensory modalities—specifically geometric depth and force)into the foundation model's core architecture\cite{yu2025forcevla,zhen2025tesseract}. Constructing such a unified, full-modality embodied foundation model, which can simultaneously synthesize environmental evolution and assimilate sparse, heterogeneous physical feedback, remains a formidable yet essential frontier.

Realizing such deep multi-modality integration is hindered by two intrinsic hurdles. First is the multimodal alignment failure arising from significant statistical heterogeneity disparities among heterogeneous signals\cite{peng2022balanced,dai2025unbiased,guo2025omnivla}.This imbalance prevents critical signals from aligning effectively within the joint distribution, suppressing their contribution to self-attention. Second is the requirement of physical consistency in multi-modal dynamics. Traditional decoupled paradigms often lead to temporal misalignment between sensory forecasting and action generation\cite{li2024multimodal}. This failure to synchronize latent imagination with motor execution compromises performance in high-precision, long-horizon tasks, necessitating a unified denoising process\cite{guo2024prediction,zitkovich2023rt} to ensure sensory-motor coherence.

To address these challenges, we propose DeMUSE, a unified diffusion transformer architecture tailored for multi-modality embodied synthesis. Inspired by the multi-sensory processing logic of the biological brain\cite{kawato2024interplay}, we treat RGB images, geometric depth, and 6-axis force as a unified serialized stream. By concatenating tokens from all modalities into a single Transformer backbone with a global receptive field, our model achieves holistic deep fusion within a unified computational flow. Inspired by recent joint denoising paradigms \cite{chi2025wow,videovla,guo2024prediction}, our architecture adopts a Joint Denoising objective to synchronously synthesize environmental evolution and action sequences.This coupling ensures that visual imagination and physical execution strictly adhere to the same causal laws of environmental.

To overcome the alignment difficulties of heterogeneous signals\cite{dit}, we design the Adaptive Modality-specific Normalization (AdaMN) mechanism and a Sparse Mixture-of-Experts (MoE) architecture with residual shared experts. AdaMN performs modality-aware feature recalibration via specialized affine expert paths for each modality. Concurrently, the MoE architecture expands model capacity while maintaining constant FLOPs, enabling the foundation model to store large-scale physical priors without compromising the low inference latency required for real-time control\cite{ditmoe}. Furthermore, addressing the lack of force feedback in existing datasets, we release an augmented MetaWorld MT50\cite{yu2020meta} benchmark and develop a robust real-world data acquisition pipeline with heterogeneous sampling alignment.

The core contributions of this work are as follows:
\begin{itemize}
    \item We present the DeMUSE model, which achieves deep synthesis of perception and action via token concatenation, employing a joint denoising paradigm to ensure physical consistency in generated sequences.
    \item We propose AdaMN to resolve numerical scale disparities and ensure multimodal alignment in multi-modality fusion. Additionally, we leverage a sparse MoE architecture for efficient capacity scaling while preserving real-time responsiveness, essential for high-precision and long-horizon embodied control.
    \item Addressing the scarcity of multimodal interaction data, we contribute an augmented MT50 benchmark featuring high-precision, synchronized 6-DoF force and depth signals. Furthermore, we establish a comprehensive real-world acquisition framework—comprising a low-latency teleoperation interface and automated synchronization tools—to generate a high-fidelity dataset for dexterous robot manipulation (Figure~\ref{real}).
\end{itemize}

\begin{figure*}[t]  
  \centering
  \includegraphics[width=0.9\textwidth]{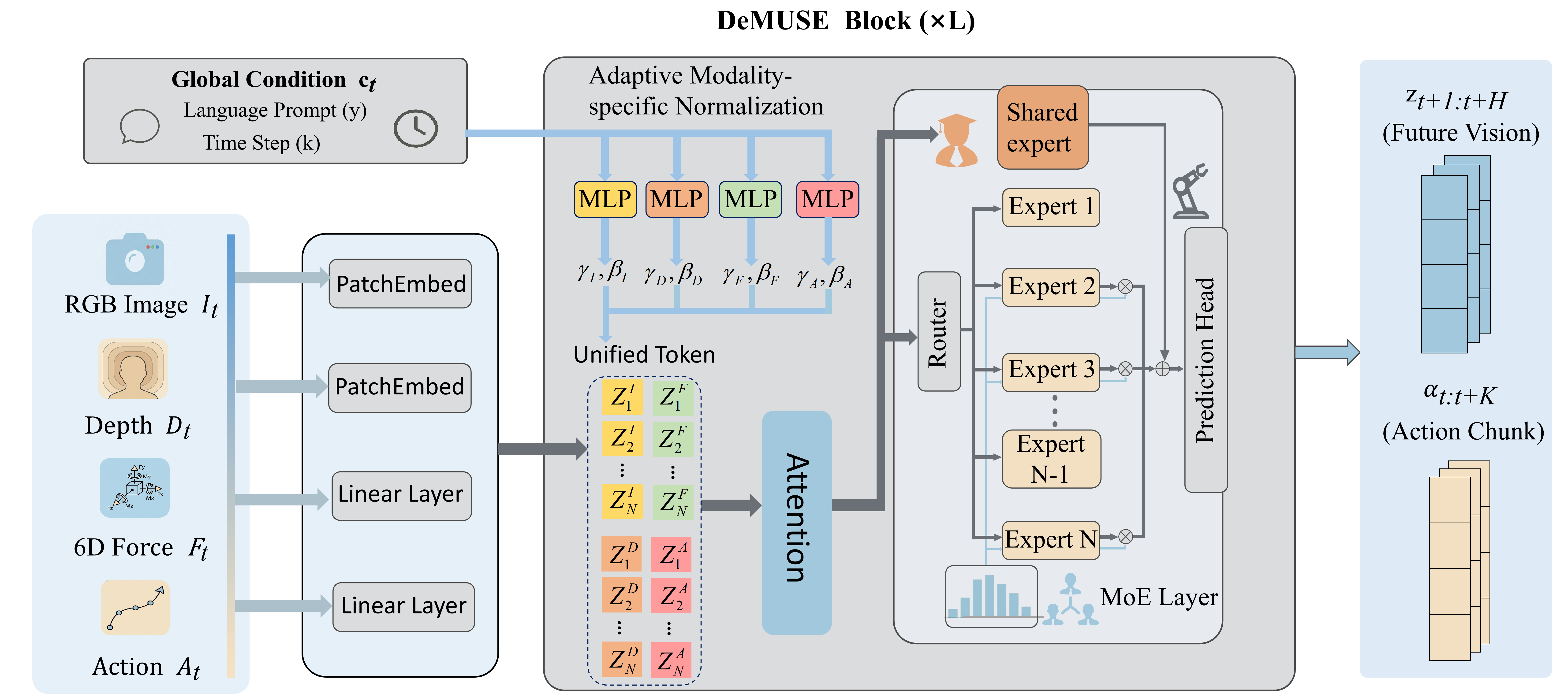}  
  \caption{\textbf{The DeMUSE architecture for multi-modal embodied synthesis.} 
Heterogeneous tokens are processed through a sequence of Transformer blocks. Each block employs AdaMN to ensure representational balance across modalities and a Sparse MoE layer to achieve high-capacity inference under real-time constraints. The framework synchronously generates future visual evolutions and continuous actions via a unified joint denoising stream.}
  \label{fig:model-framework}
  \vspace{-0.2in}
\end{figure*}

\section{Related Work}

\paragraph{Generative Models and Multimodal Foundation Models.} Diffusion-based paradigms\cite{chi2025diffusion,guo2024prediction,chi2025wow,lin2025storm,yang2025covar,intelligence2504pi0} have demonstrated exceptional expressive power in modeling complex, multimodal action distributions. The field has transitioned from CNN-based reactive policies toward Diffusion Transformers (DiT)\cite{dit,wang2025lavin} to facilitate unified processing of heterogeneous tokens and large-scale pre-training. Recent world models\cite{ha2018world,hafner2023mastering,videovla,chi2025wow} extend this by situating environmental evolution and action synthesis within a shared latent space, thereby acquiring physical common sense through predictive learning. However, these vision-centric paradigms often neglect non-visual physical anchors, failing to resolve the numerical scale disparities between high-dimensional visual streams and sparse force signals\cite{navarro2023visuo}. Recent research has also focused on expanding the perceptual modality\cite{mao2024multimodal,liurdt,yu2025forcevla,guo2025omnivla},such endeavors typically integrate depth or force data as auxiliary constraints, manifesting "cerebellar-like" reactive characteristics that prioritize immediate feedback over profound cognitive integration. Neuroscience research underscores that the Cerebrum, rather than the cerebellum, functions as the central nexus for synthesizing highly heterogeneous sensory signals—vision, tactile, and proprioception—into unified cognitive representations\cite{ernst2011multisensory}. Inspired by these insights, DeMUSE introduces a "cerebral-style" deep synthesis framework where modality tokens co-evolve within a unified denoising flow. This architecture facilitates effective interaction between extremely heterogeneous signals by resolving the multimodal alignment challenges and representational imbalances inherent in divergent information densities.

\paragraph{Efficient Scaling with Sparse Experts.} In recent years, numerous studies in the field of robotics have demonstrated the effectiveness of scaling laws in embodied intelligence and foundation models\cite{zhu2025scaling,shukor2025scaling,lindata}. However, dense scaling significantly increases per-step computational costs (FLOPs), leading to inference latencies that violate the stringent real-time constraints of robotic control. Sparse Mixture-of-Experts (MoE)\cite{sun2024ec,reuss2024efficient,cheng2025moe} addresses this by decoupling total parameter count from activated compute. By strategically routing tokens, MoE allows individual experts to specialize in distinct physical dynamics or sensory modalities\cite{shazeer2017outrageously, lepikhin2020gshard,sunec,yu2025forcevla}, thereby enhancing representational capacity. Despite these benefits, sparse architectures often suffer from expert collapse and routing instability during training\cite{deepseekmoe}. To ensure training convergence and balanced utilization, auxiliary load-balancing losses are typically employed to regularize the gating network\cite{guo2025advancing}. We utilize this sparse expansion as a foundation for deep multi-modal integration, achieving substantial performance gains while strictly preserving real-time operational efficiency.

\section{Methodology}
\subsection{Problem Formulation}

To ensure physically consistent manipulation, we adopt a multi-modality joint denoising paradigm. This framework models the co-evolution of environmental states and agent actions within a unified representation space, conditioned on multimodal sensory observations.

We consider an offline dataset $\tau = \{(o_t, a_t)\}_{t=0}^{T-1}$, where at each timestep $t$, the agent acquires heterogeneous observations $o_t = (I_t, \mathcal{D}_t, f_t)$, comprising RGB images $I_t$, spatial depth $\mathcal{D}_t$, and 6-axis force feedback $f_t$.

\textbf{Unified Tokenization and Joint Objective.} All sensory modalities are projected into a shared $D$-dimensional latent space and serialized as tokens. We define the \textit{joint objective} $\mathbf{w}_0 = [\mathbf{z}_{t+1:t+H}, \mathbf{a}_{t:t+K-1}]$, representing future visual evolution over $H$ steps and the action sequence over $K$ steps. This formulation enables the transformer to capture cross-modal dependencies between physical transitions and motor commands\cite{zhang2024multimodal}.

\textbf{Joint Denoising Process.} We adopt a latent refinement framework where the objective $\mathbf{w}_0$ is perturbed into a noisy state $\mathbf{w}_k$ over $k$ steps. The model learns a conditional denoiser $\boldsymbol{\epsilon}_\theta(\mathbf{X} \mid y, k)$ to recover the clean target from a unified token sequence $\mathbf{X} = \{z_t, d_t, f_t, \mathbf{w}_k\}$. By attending to the geometric structure and contact semantics during iterative refinement, DeMUSE enforces physical consistency across the synthesized imagination and action trajectories.

\subsection{Model Architectures}\textbf{System Components.} DeMUSE is a Diffusion Transformer optimized for omni-modal embodied synthesis, employing a unified denoising stream to synchronously generate future latent trajectories and continuous action chunks (Figure~\ref{fig:model-framework}). The Heterogeneous Embedding Layer maps observations into a unified token space via modality-specific encoders: RGB images are encoded into VAE latent space with patch-projection\cite{vae}, while depth maps, 6-axis force, and action signals are projected through dedicated linear layers. The input sequence is formed by concatenating clean context with noisy target variables for both visual and action modalities. These tokens are processed by a Multimodal Backbone comprising a stack of $L$ Transformer blocks for cross-modal global integration, followed by a Joint Prediction Head that estimates the noise $\hat{\boldsymbol{\epsilon}}$ to drive the reverse diffusion process.

\textbf{Block Design.} Within each block, a Modulate--Synthesize--Scale hierarchy (Figure~\ref{fig:model-framework}) maximizes multimodal synergy. The AdaMN mechanism recalibrates feature distributions through modality-specific affine transformations conditioned on the global context. This is followed by a bidirectional attention mechanism that facilitates deep causal synthesis between visual geometry and physical dynamics. To expand model capacity without compromising real-time control latency, a MoE layer is integrated to assimilate complex multimodal priors. This structure ensures the model can store vast physical knowledge while maintaining the fixed-frequency execution required for robotic tasks.

\textbf{Training Protocol.} We adopt a two-stage paradigm~\cite{ghosh2024octo,guo2024prediction}: large-scale pre-training on the subset of Open X-Embodiment (OXE) dataset~\cite{oxe} to distill universal physical priors, followed by domain-specific fine-tuning. This strategy ensures robust convergence and high-precision execution by adapting the model to specific observation noise and contact dynamics. Comprehensive technical details regarding the pre-training datasets, weight initialization, and hyperparameter configurations are provided in \textbf{Appendix~\ref{appendix:training}}.

\subsection{Adaptive Modality-specific Normalization}\label{sec:adamn}Despite the unified latent space, aligning RGB, depth, and force signals remains challenging due to significant statistical heterogeneity. Standard AdaLN typically homogenizes these signals by averaging statistics across heterogeneous modalities , leading to representation interference and a failure to capture distinct physical dynamics.To address this, we propose AdaMN, which performs modality-aware feature recalibration. Instead of a monolithic projection, we employ dedicated expert layers $\phi_m$ for each modality $m$. These layers dynamically synthesize modality-specific scale ($\gamma_m$) and offset ($\beta_m$) factors conditioned on the \textbf{global context} $c = \text{MLP}(y, k)$, which integrates both the language instruction $y$ and the diffusion timestep $k$:
\begin{equation}
[\gamma_m, \beta_m] = \phi_m(c), \quad h'_m = \gamma_m \cdot \hat{h}_m + \beta_m
\label{eq:adamn}
\end{equation}
Through these bifurcated mapping paths, AdaMN ensures that each modality retains its activation specificity and is not suppressed by the statistical averaging of the shared computational stream. 
Furthermore, AdaMN provides a flexible foundation for embodied models. By dynamically allocating experts based on available sensor configurations, it adaptively accommodates missing modalities without hard-coded constraints, maintaining representational balance during deep fusion.

\subsection{Scaling with Mixture-of-Experts}\label{sec:moe_scaling}To satisfy high-capacity requirements for multimodal fusion while adhering to real-time control constraints, we replace the standard Feed-Forward Network (FFN) with a sparse Mixture-of-Experts (MoE) layer featuring a  shared-expert design. Unlike conventional sparse architectures, this design incorporates a perennially active shared branch\cite{deepseekmoe,ditmoe} to stabilize the representation of heterogeneous signals.

\textbf{Routing Mechanism.} For each input token $h$, a gating network $G(h)$ determines the expert allocation. Previous research has demonstrated that such a shared structure significantly enhances training stability. The gate first projects $h$ into a logit space to compute routing scores. These scores are then normalized via a softmax function to produce a probability distribution over the available routed experts \cite{routing}.Empirical evaluations of DeMUSE indicate that top-1 and top-2 routing configurations yield comparable success rates across our benchmark tasks; detailed comparative data for these configurations are provided in \textbf{Appendix~\ref{appendix:arch}}. Consequently, to further minimize computational overhead and maximize inference speed for real-time control, we adopt top-1 routing as our default strategy. 
\begin{equation}y = E_{\text{shared}}(h) + \sum_{i \in \text{Top-}k} G(h)_i E_i(h)\label{eq:moe_output}\end{equation}
where $E_{\text{shared}}$ is the shared expert capturing invariant physical features and universal cross-modal priors, while gated experts $\{E_i\}$ specialize in modality-specific dynamics.

\textbf{Representational Stability.} The shared expert provides a consistent gradient path, mitigating the routing oscillations typically associated with fully sparse gating on heterogeneous inputs\cite{ditmoe,deepseekmoe}. By decoupling universal perceptual priors in the shared branch from complex physical interactions in the routed experts, the architecture achieves efficient parameter scaling while maintaining constant inference FLOPs. This ensures DeMUSE retains high generalization bounds without compromising the control frequency required for closed-loop manipulation.

\subsection{Learning Objectives and Inference}\label{sec:training_details}
\textbf{Joint Learning Objective.} We optimize DeMUSE using a multi-objective function designed to ensure stable physically-consistent trajectory generation. The model operates by minimizing the noise prediction error within the unified multimodal latent space. Specifically, the total loss $\mathcal{L}$ is formulated as a combination of the Mean Squared Error ($\mathcal{L}_{\text{mse}}$) for precise denoising and a Variational Lower Bound ($\mathcal{L}_{\text{vb}}$) to regularize the latent distribution:
\begin{equation}
\mathcal{L} = \lambda_v (\mathcal{L}_{\text{mse}}^v \!+\! \mathcal{L}_{\text{vb}}^v) + \lambda_a (\mathcal{L}_{\text{mse}}^a \!+\! \mathcal{L}_{\text{vb}}^a) + \alpha \mathcal{L}_{\text{aux}}
\end{equation}
where $\lambda_v$ and $\lambda_a$ are weighting coefficients for visual and action modalities, respectively. Notably, force ($f_t$) and depth ($d_t$) serve as fixed physical conditions and do not contribute to loss gradients, ensuring the model focuses on learning generative dynamics under environmental constraints.To prevent expert collapse, we employ an auxiliary load-balancing loss $\mathcal{L}_{\text{aux}}$:
\begin{equation}
\mathcal{L}_{\text{aux}} = \alpha N \textstyle \sum_{i=1}^{N} \mathbb{E}_{h \in \mathcal{B}}[g_{i}^{(h)}] \cdot \mathbb{E}_{h \in \mathcal{B}}[s_{i}^{(h)}]
\end{equation}
where $\mathcal{B}$ denotes the set of non-action tokens. By confining $\mathcal{L}_{\text{aux}}$ to dense perceptual modalities, we prevent the global load-balancing regularization from overriding the specialized expert allocation for sparse action signals, thereby preserving their distinct physical semantics.

\textbf{Inference and Action Chunking.} To ensure temporal coherence and mitigate execution jitter, we implement action chunking by synchronizing the prediction horizon $H$ with the execution stride $K$. This alignment enforces causal coupling between imagined visual evolution and motor trajectories while minimizing re-planning errors. Furthermore, closed-loop control employs Exponential Moving Average (EMA) parameters to act as a temporal filter, suppressing stochastic noise for smooth manipulation. This protocol effectively synergizes long-range hallucination with real-time corrections for robust, precise control.

\begin{table*}[t]
\centering
\caption{\textbf{Real-World Task Suite and Semantic Instructions.} We design four representative tasks in unstructured environments to evaluate embodied manipulation. These tasks present significant physical and coordination challenges, involving intricate constraints such as multi-stage contact and high-precision spatial alignment across diverse linguistic commands.}
\label{tab:real_world_tasks}
\small
\renewcommand{\arraystretch}{1.3}
\newcolumntype{L}{>{\RaggedRight\arraybackslash}X} 

\begin{tabularx}{\textwidth}{l l L}
\toprule
\textbf{Task Name} & \textbf{Core Physical Challenge} & \textbf{Language Instruction ($y$)} \\
\midrule
Tool Drawer Org.    & Long-duration manipulation & Pick up the tool from the table, place it in the open drawer, and close all drawers. \\
Fill Cup (1/3 Cola) & Precise control            & Grab the cup from the table, then fill it with one-third of cola from the dispenser. \\
Dispense Sanitizer  & Pressure sensitivity       & Dispense a suitable amount of hand sanitizer onto one hand from the press bottle. \\
Sweeping to Dustpan & Visual occlusion           & Use the broom to sweep the scattered items from the table into the dustpan. \\
\bottomrule
\end{tabularx}
\end{table*}

\begin{table*}[t]
\centering
\caption{\textbf{Main Results on MetaWorld MT50.} We report the success rate ($\%$) averaged over 50 evaluation trials per task. The representative tasks highlight DeMUSE's precision in diverse scenarios.}
\label{tab:main_results_streamlined}
\small
\begin{tabularx}{\textwidth}{l @{\extracolsep{\fill}} ccc cccc | c}
\toprule
\multirow{2}{*}{\textbf{Method}} & \multicolumn{3}{c}{\textbf{Success Rate by Difficulty ($\%$)}} & \multicolumn{4}{c}{\textbf{Representative Tasks (SR $\%$)}} & \multirow{2}{*}{\shortstack{\textbf{Avg. SR} \\ \textbf{(\%)}}} \\
\cmidrule(lr){2-4} \cmidrule(lr){5-8}
& Simple & Medium & Hard & \textit{Hammer} & \textit{Basket} & \textit{Dial} & \textit{Plate-slide} & \\
\midrule
DP             & 45.5 & 27.4 & 15.3 & 8.0 & 10.0 & 52.0 & 60.0 & 29.2 \\
RT-2           & 81.3 & 50.4 & 25.6 & 20.0 & 12.0 & 52.0 & 88.0 & 52.2 \\
PAD            & 92.3 & 65.8 & 61.3 & 82.0 & 84.0 & 56.0 & 72.0 & 72.4 \\
ForceVLA       & 95.6 & 71.4 & 62.1 & 88.0 & 85.0 & 82.0 & 94.0 & 75.9 \\
RDT-1B         & 94.2 & 76.7 & 63.3 & 86.0 & 89.0 & 78.0 & 92.0 & 77.9 \\
\midrule
\textbf{DeMUSE (Ours)} & \textbf{95.7} & \textbf{83.8} & \textbf{69.8} & \textbf{96.0} & \textbf{96.0} & \textbf{96.0} & \textbf{100.0} & \textbf{83.2} \\
\bottomrule
\end{tabularx}
\end{table*}

\section{Experiments}
\label{sec:experiments}
In this section, we conduct a systematic experimental evaluation to verify DeMUSE's effectiveness in handling complex embodied manipulation tasks. Our experimental design is structured to address the following core research questions:

 \textbf{Q1 (Comprehensive Performance):} Does DeMUSE significantly outperform existing state-of-the-art Vision-Language-Action (VLA) models and diffusion-based policy baselines in multi-task embodied manipulation scenarios?  \textbf{Q2 (Multimodal Synergy):} How does the deep fusion of heterogeneous perceptual modalities (vision, depth, and force) enhance decision-making robustness in precision contact tasks and unstructured environments? \textbf{Q3 (Architecture Ablation):} What is the specific role of the AdaMN modulation mechanism in maintaining the balance of multimodal representations? \textbf{Q4 (Scaling and Efficiency):} Can the MoE architecture, featuring a Residual Shared-Expert design, elevate the model's performance upper bound through parameter scaling while maintaining the inference efficiency requisite for real-time robotic control?  

\begin{figure*}[t]
  \centering
  \includegraphics[width=0.95\textwidth]{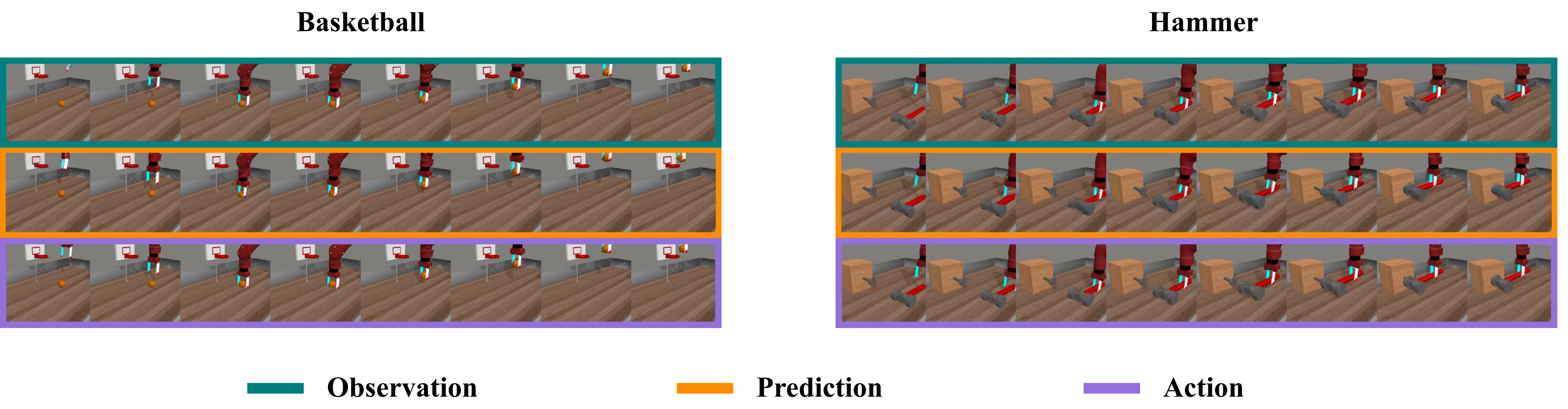} 
  \caption{\textbf{Qualitative results on MetaWorld.} Left: Basketball task; Right: \textsc{Hammer} task. Rows from top to bottom represent input Observations, generated Predictions, and actual Actions. High consistency across rows demonstrates the model's ability to capture physically-consistent latent dynamics through joint denoising.}
  \vspace{-0.2in}
  \label{fig:meta_visualization}
\end{figure*}

\subsection{Datasets and Baselines}
\noindent\textbf{Enhanced Heterogeneous Simulation.} We developed an augmented MetaWorld MT50 benchmark \cite{yu2020meta} by integrating high-precision, synchronized 6-DoF force and depth signals to support dexterous robot manipulation. While the original framework provides a diverse task suite, we significantly enhanced the environment to facilitate deep multi-sensory fusion. To ensure physical consistency, force signals are meticulously transformed into the end-effector (EE) local frame, refined via Savitzky-Golay smoothing, and normalized. This enhanced platform provides a rigorous and high-fidelity baseline for evaluating omni-modal policies under complex physical interaction.

\noindent\textbf{Real-world Platform and Data Pipeline.} Our hardware suite comprises a UR10 robotic arm, an OnRobot RG2-FT force-controlled gripper, and an Intel RealSense D435i camera. We developed a robust data pipeline utilizing an event-driven asynchronous architecture and dual-timestamping to synchronize heterogeneous sensor rates and mitigate clock drift. To ensure balanced representation, we collected 100 expert trajectories for each of the four tasks detailed in Table~\ref{tab:real_world_tasks} (totaling 400 trajectories). This platform validates the deep fusion and closed-loop robustness of DeMUSE for dexterous robot manipulation in unstructured environments.

\textbf{Baselines and Experimental Fairness.} To ensure a rigorous and impartial quantitative assessment, all models are trained on identical datasets using a consistent fine-tuning protocol. A defining characteristic of \texttt{DeMUSE} is its native capacity for deep omni-sensory synthesis, which seamlessly integrates RGB, depth, and 6-axis force into a unified serialized stream. To maintain experimental parity, we provide each baseline with the maximal sensory information its architecture is natively designed to support—ensuring that RGB and proprioceptive inputs are standardized across all models, while providing high-fidelity depth and force feedback to baselines with corresponding support. This controlled setup ensures that the observed performance gains are directly attributable to the policy’s representation efficiency and its ability to reason across heterogeneous modalities. We compare DeMUSE against five state-of-the-art baselines: \textbf{(1) Diffusion Policy}~\cite{chi2023diffusion}: A representative action-only diffusion paradigm that performs iterative denoising conditioned on visual observations; 
\textbf{(2) RT-2}~\cite{zitkovich2023rt}: A landmark Vision-Language-Action (VLA) model utilizing discrete action tokenization to facilitate internet-scale semantic transfer; 
\textbf{(3) PAD}~\cite{guo2024prediction}: A DiT-based policy focused on joint video-action synthesis to capture complex environment dynamics; 
\textbf{(4) ForceVLA}~\cite{yu2025forcevla}: An extension of the $\pi_0$ architecture that integrates a flow-matching mechanism with real-time 6-axis force feedback; 
\textbf{(5) RDT-1B}~\cite{liurdt}: A 1.2B-parameter diffusion transformer featuring a physically interpretable unified action space designed to capture the non-linear and high-frequency dynamics of heterogeneous robot data.

\subsection{Results and Analysis} \label{sec:results_analysis}

\noindent\textbf{Fine-tuning Protocol and Multi-task Performance.} We fine-tune DeMUSE on the MT50 multi-task benchmark using 4 NVIDIA A100 GPUs. For each task, 50 expert trajectories are utilized for training. The model is optimized for 100,000 gradient steps with a global batch size of 64.

Table \ref{tab:main_results_streamlined} reports the mean success rates (\%) across all 50 tasks. The results demonstrate that DeMUSE achieves the highest overall success rate of 83.2\%, outperforming the strongest baselines in Medium and Hard tasks by 7.1 and 6.5 percentage points, respectively. In contact-intensive tasks such as \textsc{Hammer-v2} and \textsc{Basketball-v2}, success rates reach 96.0\%, indicating that omni-modal deep fusion allows the model to effectively capture critical contact dynamics.

To qualitatively analyze the generative control logic, Figure \ref{fig:meta_visualization} visualizes the joint denoising process on two representative tasks. The rows sequentially present the observations, the model's predicted future visual evolution, and the actual execution results. The high alignment between synthesized futures and physical rollouts demonstrates that DeMUSE has internalized physically-consistent latent dynamics, enabling precise intervention during complex interactions.
\begin{table}[h]
\centering
\caption{\textbf{Real-World Robot Success Rates (\%).} Comparison of DeMUSE against state-of-the-art baselines across four representative physical tasks. Results are averaged over 50 independent trials per task with randomized initializations.}
\label{tab:real_world_results}
\small
\setlength{\tabcolsep}{8pt}
\begin{tabular}{lccc}
\toprule
\textbf{Task Name}           & \textbf{DP} & \textbf{RDT-1B} & \textbf{Ours} \\
\midrule
Sweeping              & 54.0        & 62.0            & \textbf{76.0} \\
Dispense Sanitizer    & 42.0        & 58.0            & \textbf{74.0} \\
Tool Drawer Org.      & 0.0        & 64.0            & \textbf{72.0} \\
Fill Cup              & 10.0        & 52.0            & \textbf{68.0} \\
\midrule
\textbf{Overall Average}      & 26.5        & 59.0            & \textbf{72.5} \\
\bottomrule
\vspace{-0.4in}
\end{tabular}
\end{table}

\noindent\textbf{Experimental Protocol and Results.} All models were fine-tuned on a shared expert dataset to ensure parity. Due to high deployment overhead, we prioritized two representative baselines: Diffusion Policy (DP) and RDT-1B. Each task involved 50 trials with randomized initial poses and unstructured disturbances. DeMUSE achieved a 72.5\% success rate in physical environments, outperforming RDT-1B and DP by bridging the sim-to-real gap through exceptional robustness in complex real-world settings. We further analyze these systemic advantages across two dimensions:

\begin{figure}[ht]
  \begin{center}
    \centerline{\includegraphics[width=\columnwidth]{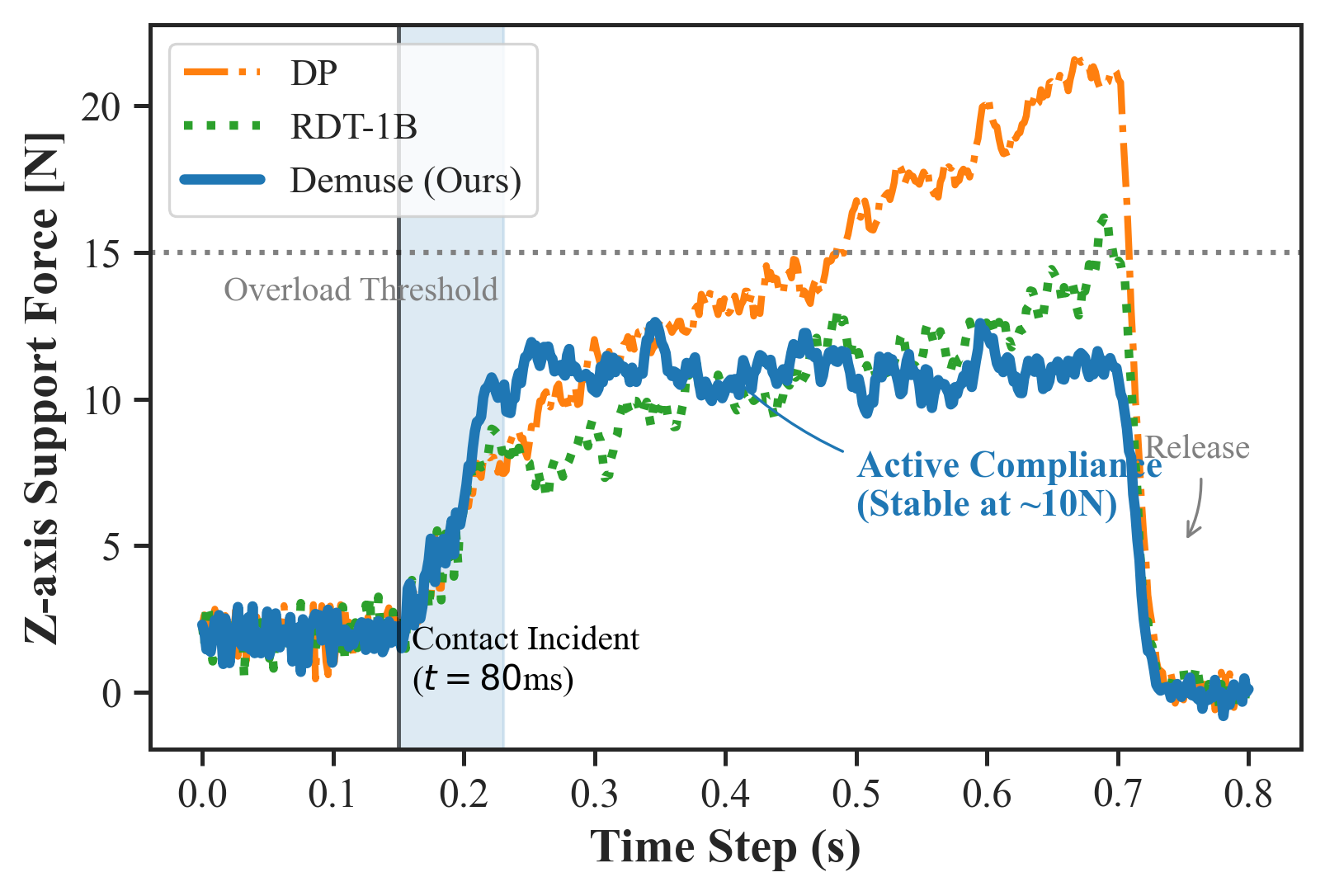}}
    \caption{
      \textbf{Z-axis force profile.} Comparison of contact dynamics during the pressing task. DeMUSE achieves stable active compliance ($\sim$10N) within 80ms of contact ($t=80$ms)
    }
    \vspace{-0.5in}
    \label{fig:force_curve}
  \end{center}
\end{figure}

\noindent \textbf{(1) Contact Dynamics and Active Compliance.} We evaluate force responsiveness using the Dispense Sanitizer task as a representative case. As shown in Figure~\ref{fig:force_curve}, DeMUSE triggers action corrections within $\sim$80ms of contact, maintaining stable compliance ($\sim$10N) while baselines suffer from hazardous force surges ($>$15N). This responsiveness is driven by the AdaMN mechanism, which recalibrates heterogeneous feature scales to prioritize sparse force signals during physical interaction, ensuring closed-loop stability where visual-only feedback fails.

\begin{figure}[ht]
  \begin{center}
    \centerline{\includegraphics[width=0.9\columnwidth]{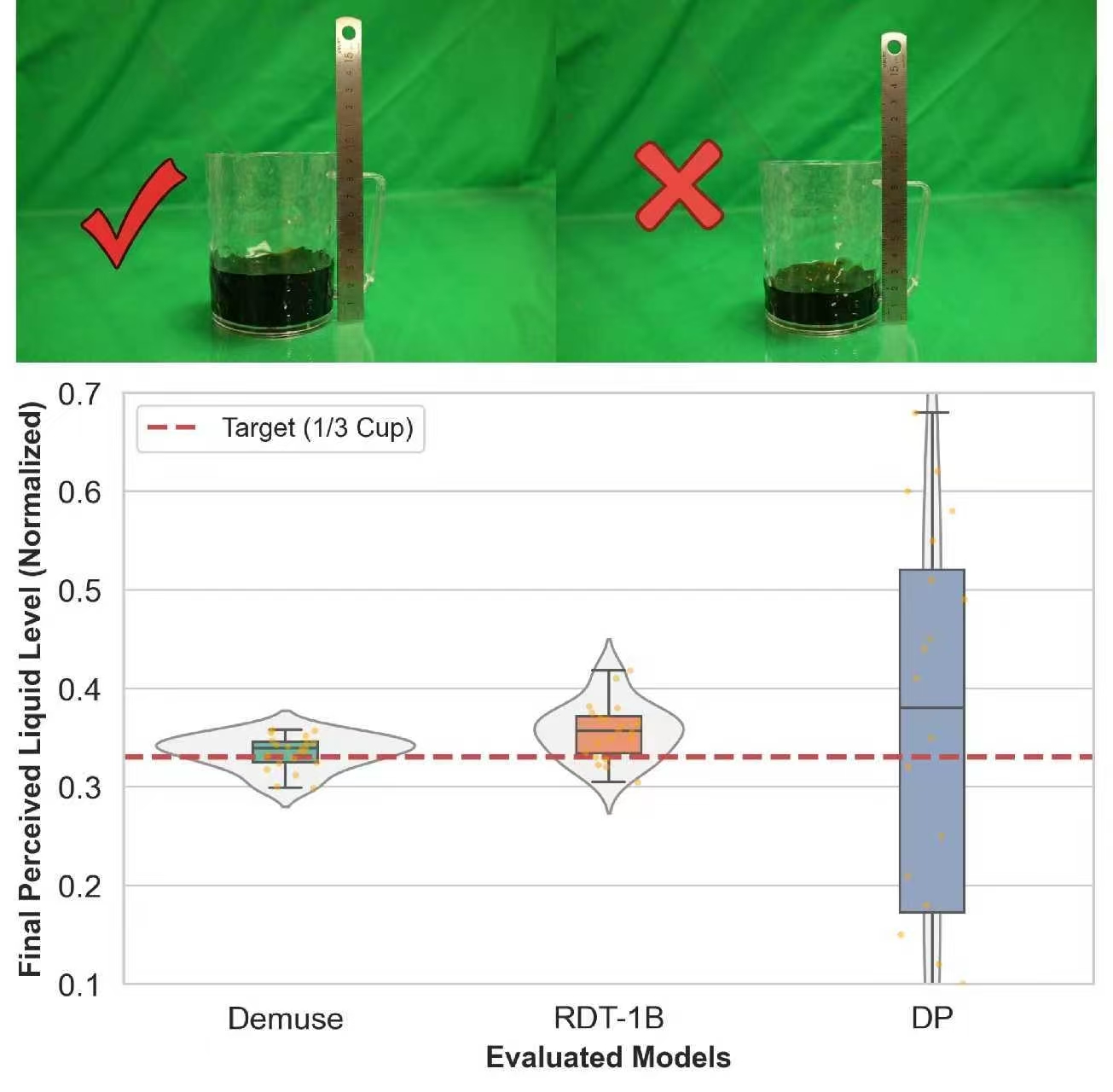}}
    \caption{
      \textbf{Terminal liquid level precision.} Final liquid level distributions in the "Fill Cup" task. DeMUSE achieves high-precision ($0.33 \pm 0.02$) with minimal variance.
    }
    \vspace{-0.3in}
    \label{fig:liquid_level}
  \end{center}
\end{figure}

\noindent \textbf{(2) Terminal State Precision in Long-Horizon Tasks.}
DeMUSE exhibits remarkable decision coherence during multi-stage transitions in tasks like "Tool Drawer Org." and "Fill Cup." In the precision-critical filling task (Figure~ \ref{fig:liquid_level}), the model achieves a terminal accuracy of $0.33 \pm 0.02$. Unlike baselines that suffer from temporal response lag, our omni-modal deep fusion leverages the synergy between vision and 6D force to enhance the model’s anticipatory capacity for physical evolution, ensuring precise system shutdown within complex dynamic environments.

\subsection{Ablation Studies}
\label{sec:ablations}

We conduct ablations to validate DeMUSE, addressing: (i) Multimodal Synergy under physical perturbations; (ii) Fusion Strategy (deep vs. conditional modulation); (iii) Modality Alignment via AdaMN; and (iv) Scaling Efficiency using sparse MoE for real-time control.

\noindent\textbf{Significance of Multimodal Integration.} We first quantify the contribution of individual sensory modalities to the decision-making process. As summarized in Table \ref{tab:ablation_modality}, force signals (RGB-F) prove pivotal for high-precision dexterous manipulation, yielding a 9.0\% improvement in success rate for "Hard" tasks in MetaWorld. Notably, while depth integration provides marginal gains in simulation—potentially due to the inherent stability of synthetic RGB signals—its contribution becomes significantly more pronounced in real-world trials, where it provides essential geometric grounding against sensory noise and environmental uncertainty.

\begin{table}[h]
\centering
\caption{\textbf{Ablation of Modality Configurations and Fusion Strategies (Success Rate \%).} 
Full fusion (Ours) significantly outperforms both modality-restricted baselines and the conditional modulation strategy (DeMUSE-Conditioned), proving that the integration architecture is as critical as the sensory data itself.}
\label{tab:ablation_modality}
\small
\addtolength{\tabcolsep}{-2pt} 
\begin{tabular}{lcccc}
\toprule
\multirow{2}{*}{\textbf{Configuration}} & \multicolumn{3}{c}{\textbf{Sim. (MetaWorld)}} & \textbf{Real-World} \\
\cmidrule(r){2-4} \cmidrule(l){5-5}
 & Simple & Medium & Hard & (Unstructured) \\ 
\midrule
RGB-only             & 93.6   & 77.3   & 60.8 & 45.5 \\
RGB-D                & 93.5   & 77.6   & 61.2 & 61.4 \\
RGB-F                & 94.7   & 81.8   & 68.2 & 58.0 \\ 
\midrule
Conditioned    & 93.8   & 77.5   & 61.1 & --   \\ 
\textbf{Ours (Full)} & \textbf{95.7} & \textbf{83.8} & \textbf{69.8} & \textbf{72.5} \\ 
\bottomrule
\end{tabular}
\vspace{-0.1in}
\end{table}

In real-world deployments, the vision-only policy (RGB-only) suffers from substantial performance degradation (-27.0\%) due to environmental noise. In contrast, DeMUSE effectively mitigates perceptual distribution shifts by leveraging stable geometric constraints from depth and physical interaction ground-truth from 6-axis force feedback. This multi-sensory integration ensures cross-scenario action consistency, particularly when visual signals are compromised.

\noindent\textbf{Deep Fusion vs. AdaMN-Conditioning.} To justify serializing physical signals into a unified stream, we compare DeMUSE with DeMUSE-Conditioned. In this variant, depth and force are excluded from the self-attention sequence; instead, they are injected into the transformer blocks exclusively via the AdaMN as modulation parameters, treating physical feedback as a global latent condition.As shown in Table \ref{tab:ablation_modality}, the performance of DeMUSE-Conditioned is nearly identical to the RGB-only baseline. This parity demonstrates that mere feature modulation fails to resolve the complex temporal correlations between sensory signals and motor commands. Only the deep fusion paradigm enables modality tokens to co-evolve within a global receptive field , which is indispensable for internalizing consistent physical laws during dexterous interactions.

\noindent\textbf{Ablation of AdaMN in Modality Alignment.} We evaluate AdaMN against Shared AdaLN using LPIPS (trajectory-averaged perceptual similarity between predictions and ground truth) and Success Rate (SR). As shown in Table~\ref{tab:adamn}, AdaMN achieves superior LPIPS (0.35 vs. 0.41), demonstrating a more precise world-modeling of environmental evolution. The higher SR (83.2\% vs. 78.6\%) further validates the accurate modulation of action modalities. This gain stems from the modality-specific affine recalibration, which decouples high-dimensional vision from sparse physical signals. In contrast, Standard AdaLN yields a compromised statistical average, failing to capture the distinct dynamics of individual modalities. These results confirm that AdaMN is essential for stabilizing the optimization of multi-modality transformers, ensuring that force and geometric cues are preserved during joint denoising.

\begin{table}[h]
\centering
\caption{\small \textbf{Normalization Ablation.} AdaMN ensures both higher world-modeling fidelity and execution precision.}
\label{tab:adamn}
\small
\begin{tabular}{lcc}
\toprule
\textbf{Method} & \textbf{LPIPS} ($\downarrow$) & \textbf{Success Rate} ($\uparrow$) \\ 
\midrule
Standard AdaLN & 0.41 &78.6\% \\ 
\textbf{AdaMN (Ours)} & \textbf{0.35} & \textbf{83.2\%} \\ 
\bottomrule
\end{tabular}
\vspace{-0.1in}
\end{table}

\begin{figure}[ht]
  \begin{center}
    \centerline{\includegraphics[width=\columnwidth]{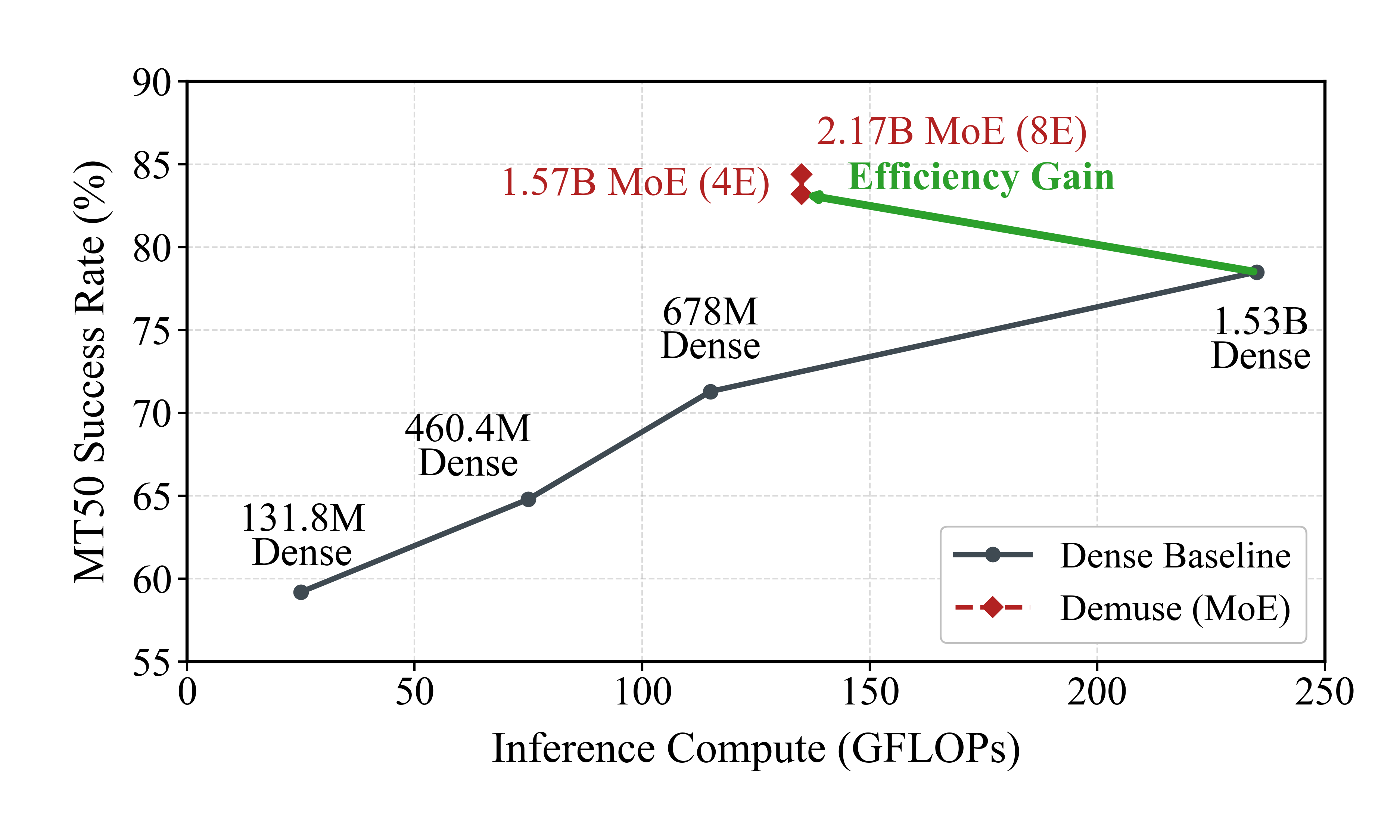}}
    \caption{
      \textbf{Scaling Analysis on MetaWorld MT50.} Comparison of model capacity (parameters), inference compute (GFLOPs), and average success rate across model variants. }

    \label{fig:scalling}
  \end{center}
  \vspace{-0.5in}
\end{figure}

\noindent\textbf{Scaling Analysis and MoE Specialization.} We investigate scaling trajectories by comparing dense baselines (131.8M--1.53B) against sparse MoE variants (1.57B/4E and 2.17B/8E). As shown in Fig.~\ref{fig:scalling}, dense models exhibit diminishing returns; the 1.53B variant attains a 78.5\% success rate but requires a prohibitive 235 GFLOPs. In contrast, MoE-4E reaching 83.2\% success with only 135 GFLOPs---a 42.6\% compute reduction. We attribute this non-linear gain to MoE-induced modality-wise specialization. Unlike dense architectures prone to feature interference, MoE allows experts to autonomously disentangle physical contact dynamics (force) from spatial geometry (vision), facilitating more robust causal reasoning in deep fusion scenarios.

\noindent\textbf{Scaling Limits and Optimal Configuration.} While extending to MoE-8E yields marginal gains (84.7\%), it introduces elevated convergence risks such as expert collapse. Given the saturation of returns and the stringent requirements for training stability, we identify MoE-4E as the optimal default. This configuration leverages expanded parameter capacity via sparse activation, sustaining robust success rates while ensuring the low-latency inference required for real-time embodied control.

\section{Conclusion} \label{sec:conclusion}

In this paper, we presented DeMUSE, a unified diffusion transformer that achieves deep multi-sensory fusion for embodied control. By integrating RGB, depth, and 6-axis force via a joint denoising objective, DeMUSE ensures physical consistency between latent imagination and action execution. We employ Adaptive Modality-specific Normalization (AdaMN) to resolve the heterogeneous distributions of multi-sensory signals and utilize a Sparse Mixture-of-Experts (MoE) architecture to enable efficient capacity scaling without compromising real-time control latency . Empirical results demonstrate significant performance gains in dexterous manipulation , validating the necessity of deep multi-modal fusion. These findings underscore that synchronized perception-action synthesis is a critical frontier for building robust, general-purpose embodied foundation models across complex and unstructured environments.

\section*{Impact Statement}
This paper presents work whose goal is to advance the field of Machine
Learning. There are many potential societal consequences of our work, none
which we feel must be specifically highlighted here.


\bibliography{main}
\bibliographystyle{icml2026}

\appendix
\onecolumn
\section{Appendix}
\label{appendix:arch}

\subsection{Heterogeneous Embedding Layer}

In the DeMUSE framework, observations from diverse modalities are projected into a unified latent space of dimension $D=1152$, corresponding to the global hidden dimension of the Transformer backbone. To ensure physical consistency during generation, we adopt a Joint Denoising embedding strategy. The specific processing logic for each modality is detailed as follows:

\begin{itemize}
    \item \textbf{RGB Images ($I_t$)}: The visual input consists of the current context frame concatenated with $H=3$ future frames to be predicted in the latent space. These frames are first compressed via a pre-trained VAE encoder, achieving an $8\times$ spatial downsampling to produce a $32 \times 32 \times 4$ latent representation. During training, Gaussian noise is applied exclusively to the future frames, while the current frame serves as a clean \textit{contextual guidance}. Subsequently, the stacked latents are processed by a projection layer with a \texttt{patch\_size} of 2, partitioning the feature map into $(32/2)^2 = 256$ non-overlapping patches. This yields an RGB token sequence of shape $[B, 256, 1152]$, supplemented with learnable positional embeddings to preserve spatio-temporal structural information.

    \item \textbf{Depth Maps ($\mathcal{D}_t$)}: Depth maps are processed at a latent resolution of $32 \times 32 \times 1$. To preserve critical geometric constraints while minimizing sequence length, we employ an independent patch embedding layer with a \texttt{d\_patch\_size} of 8. This results in a compact sequence of $(32/8)^2 = 16$ tokens with shape $[B, 16, 1152]$. To ensure precise spatial alignment between visual and geometric features, the depth positional embeddings are derived by spatially downsampling the RGB positional embeddings by a factor of 4.

    \item \textbf{6-axis Force-Torque ($f_t$)}: The haptic signal is a 6-dimensional vector comprising three-axis forces and three-axis torques in the end-effector coordinate frame. This signal is mapped to a single haptic token of shape $[B, 1, 1152]$ via a linear projection layer (\texttt{nn.Linear(6, 1152)}). This token provides high-fidelity tactile feedback and participates in global attention calculation with fixed sinusoidal positional encodings.

    \item \textbf{Actions ($A_t$)}: Consistent with the joint denoising paradigm, actions are treated as a strong physical prior. The input vector (dimension 16) is formed by concatenating the current absolute proprioceptive state with the future action sequence (corresponding to 4 continuous steps under the MetaWorld 4-DOF configuration, i.e., 1 current state and 3 future actions). Only the future action components are subjected to the diffusion noise. This vector is projected into a single conditional token of shape $[B, 1, 1152]$, enabling the model to efficiently synthesize environmental evolution conditioned on known action states.
\end{itemize}

During the forward pass, the aforementioned 274 tokens (256 RGB + 16 Depth + 1 Force + 1 Action) are concatenated in a fixed canonical order. Each token is assigned a unique \texttt{modality\_id} to facilitate modality-aware feature calibration within the subsequent AdaMN modules.

\subsection{Implementation Details of Adaptive Modality Normalization (AdaMN)}
\label{appendix:adamn}

DeMUSE introduces an Adaptive Modality Normalization (AdaMN) mechanism, which achieves deep perceptual fusion by assigning independent affine paths to each modality. The specific implementation details of this technology are as follows:

\begin{itemize}
    \item \textbf{Mathematical Formulation and Calibration}: For a feature vector $\mathbf{h}_m^{(k)} \in \mathbb{R}^{D}$ belonging to modality $m$ within the $k$-th Transformer block, AdaMN first performs an affine-free normalization, followed by a modality-specific affine transformation and diffusion-step modulation, as detailed in Algorithm \ref{alg:adamn}.
    
    \begin{algorithm}[H]
    \caption{Adaptive Modality-Normalization (AdaMN)}
    \label{alg:adamn}
    \begin{algorithmic}[1]
    \STATE {\bf Input:} Features $\mathbf{h} \in \mathbb{R}^{B \times N \times D}$, modality IDs $\mathbf{m}$, conditioning $\mathbf{e}_k \in \mathbb{R}^{B \times D}$
    \STATE {\bf Output:} Calibrated features $\mathbf{h}'$
    \STATE $(\Delta\gamma, \Delta\beta) \leftarrow \text{MLP}(\mathbf{e}_k)$ \COMMENT{Generate adaptive modulation parameters}
    \FOR{each modality $m \in \{0, \dots, M-1\}$}
        \STATE $\mathcal{I}_m \leftarrow \{i \mid \mathbf{m}_i = m\}$ \COMMENT{Group tokens by modality}
        \STATE $\tilde{\mathbf{h}}_{\mathcal{I}_m} \leftarrow \gamma_m \odot \text{LN}(\mathbf{h}_{\mathcal{I}_m}) + \beta_m$ \COMMENT{Modality-specific affine calibration}
        \STATE $\mathbf{h}'_{\mathcal{I}_m} \leftarrow (1 + \Delta\gamma) \odot \tilde{\mathbf{h}}_{\mathcal{I}_m} + \Delta\beta$ \COMMENT{Diffusion step modulation}
    \ENDFOR
    \STATE \textbf{return} $\mathbf{h}'$
    \end{algorithmic}
    \end{algorithm}

    This architecture ensures that even when statistical distributions vary significantly across modalities—particularly for sparse haptic tokens—the model preserves their \textit{activation specificity} without being suppressed by dominant visual features.

    \item \textbf{Conditioned Modulation}:
The modality-specific affine parameters $(\gamma_m,\beta_m)$ are conditioned on a global context embedding
$\mathbf{c}=\mathrm{MLP}(y,k)\in\mathbb{R}^{D}$, which integrates both the (encoded) language instruction $y$
and the diffusion timestep $k$.
Concretely, a lightweight modality expert $\phi_m$ (a 2-layer MLP with SiLU) maps $\mathbf{c}$ to two
$D$-dimensional vectors:
\[
(\gamma_m,\beta_m)=\phi_m(\mathbf{c}),\qquad \gamma_m,\beta_m\in\mathbb{R}^{D}.
\]
This design enables synchronized, condition-aware cross-modal calibration throughout the denoising trajectory.

    \item \textbf{Dimensionality Alignment}: All affine parameters are strictly aligned with the global hidden dimension $D=1152$. This consistency allows for highly efficient element-wise operations across the unified sequence $\mathbf{X} \in \mathbb{R}^{274 \times 1152}$, ensuring that the modality-aware gating mechanism introduces negligible computational overhead during inference.
\end{itemize}

\subsection{Sparse MoE Architecture with Residual Shared Experts}
\label{appendix:moe}

To scale model capacity while strictly adhering to the stringent inference latency requirements of real-time robotic control, DeMUSE replaces the conventional Feed-Forward Networks (FFNs) with sparse Mixture-of-Experts (MoE) blocks featuring a Residual Shared Expert design.

\begin{itemize}
    \item \textbf{Residual Shared Design}: Departing from purely sparse architectures, we introduce a persistently activated shared expert branch $E_{\text{shared}}$. The comprehensive forward pass logic, including routing and residue synthesis, is defined in Algorithm \ref{alg:moe}. The intermediate dimension of the shared expert is set to $4D = 4608$, equivalent to the FFN width in a dense model. This component is responsible for encoding universal physical priors, while the routed experts capture complex, modality-specific dynamics through sparse activation.

    \begin{algorithm}[H]
    \caption{DeMUSE Sparse MoE Forward Pass}
    \label{alg:moe}
    \begin{algorithmic}[1]
    \STATE {\bf Input:} Hidden states $\mathbf{x} \in \mathbb{R}^{L \times D}$, modality IDs $\mathbf{m} \in \mathbb{R}^L$
    \STATE {\bf Output:} Output states $\mathbf{y}$, auxiliary loss $\mathcal{L}_{\text{aux}}$
    \STATE \COMMENT{--- Top-1 Routing ---}
    \STATE $\mathbf{S} = \mathbf{x} \mathbf{W}_{gate}^\top$ \COMMENT{Compute raw routing logits}
    \STATE $idx, prob = \text{Top1}(\text{Softmax}(\mathbf{S}))$ \COMMENT{Select expert per token}
    \STATE \COMMENT{--- Dual-Path Computation ---}
    \STATE $\mathbf{y}_{shared} = \text{SharedFFN}(\mathbf{x})$ \COMMENT{Global knowledge via shared expert}
    \STATE $\mathbf{y}_{routed} = prob \cdot \text{Expert}_{idx}(\mathbf{x})$ \COMMENT{Specialized knowledge via routed expert}
    \STATE $\mathbf{y} = \mathbf{y}_{shared} + \mathbf{y}_{routed}$ \COMMENT{Residual expert synthesis}
    \STATE \COMMENT{--- Constrained Load Balancing ---}
    \STATE $\mathcal{X}_{aux} = \{x_i \mid \mathbf{m}_i \neq 1\}$ \COMMENT{Exclude Action tokens from balancing}
    \STATE $\mathcal{L}_{\text{aux}} = \text{ComputeLoss}(\mathcal{X}_{aux}, \mathbf{S})$ \COMMENT{Balance routing for other modalities}
    \STATE \textbf{return} $\mathbf{y}, \mathcal{L}_{\text{aux}}$
    \end{algorithmic}
    \end{algorithm}

    \item \textbf{Top-1 Routing and Performance Trade-offs}: To maximize inference efficiency and satisfy the ultra-low latency demands of embodied control, we employ a Top-1 routing strategy by default. Empirical evaluations (see Table \ref{tab:topk_comparison}) demonstrate that Top-1 and Top-2 configurations achieve comparable success rates on benchmark tasks. However, Top-1 routing significantly reduces computational overhead and enhances throughput. Specifically, the Top-1 configuration (e.g., MoE-4E) achieves an 83.2\% success rate while constraining the computational burden to 135 GFLOPs.

\begin{table}[h]
\centering
\caption{Performance Comparison of Top-1 vs. Top-2 Routing Strategies (on MetaWorld MT50)}
\label{tab:topk_comparison}
\begin{tabular}{lcc}
\hline
\textbf{Routing Strategy} & \textbf{Avg. Success Rate (\%)} & \textbf{Inference Cost (GFLOPs)} \\ \hline
Top-1 (DeMUSE) & 83.2\%  & 135  \\
Top-2           & 82.5\%  & 174  \\ \hline
\end{tabular}
\end{table}

    \item \textbf{Auxiliary Load Balance Loss}: To mitigate expert collapse during training, we incorporate an auxiliary loss $\mathcal{L}_{\text{aux}}$ with a penalty coefficient $\alpha = 0.01$. Critically, we exclude Action Tokens (\texttt{modality\_id}=1) from the load balancing statistics. This design is motivated by the extreme sensitivity of the action modality to control precision; by exempting it from global balancing regularization, we prevent action tokens from being forcibly assigned to non-specialized experts.
\end{itemize}

This "Dense-Base + Sparse-Expansion" design allows DeMUSE to scale to 1.57B parameters while maintaining a computational cost of approximately 135 GFLOPs. Compared to a dense counterpart, this architecture reduces inference power consumption by 42.6\%.

\section{Training and Implementation Details}
\label{appendix:training}

\subsection{Multi-Stage Training Strategy}

The training of \texttt{DeMUSE} follows a systematic two-stage pipeline: large-scale multi-modal pre-training followed by task-specific fine-tuning. This strategy ensures that the model internalizes broad physical priors before adapting to complex manipulation constraints.

\paragraph{Pre-training with Pure Visual Priors} 
To endow the model with robust visual and physical common sense, we perform pre-training on the Bridge V2 subset of the Open X-Embodiment (OXE) dataset. The backbone is initialized using pre-trained weights from \textbf{DiT-XL/2}, providing a strong pure visual prior for the unified latent space. For the MoE layers, we implement a \textit{sparse upcycling} protocol to transition from the dense backbone:
\begin{itemize}
    \item \textbf{Shared Expert}: This is directly initialized with the weights from the original FFN layers of the DiT-XL/2 backbone.
    \item \textbf{Routed Experts}: To break symmetry and facilitate specialization, these experts are initialized with the original FFN weights supplemented by a stochastic Gaussian perturbation.
\end{itemize}
The pre-training phase was conducted on a cluster of 8$\times$ NVIDIA A100 (80GB) GPUs for a duration of 3 days.

\subsection{Fine-tuning Configuration Summary}

Table \ref{tab:hyperparams} summarizes the key hyperparameters specifically employed during the \textbf{task-specific fine-tuning phase} of \texttt{DeMUSE}. To maintain high computational throughput during this stage, we utilized the TF32 (TensorFloat-32) precision mode on the 4-GPU cluster.

\begin{table}[h]
\centering
\caption{Hyperparameters for DeMUSE Fine-tuning Phase}
\label{tab:hyperparams}
\begin{small}
\begin{tabular}{lc|lc}
\hline
\textbf{Architecture} & \textbf{Value} & \textbf{Optimization} & \textbf{Value} \\ \hline
Transformer Layers ($L$) & 28 & Optimizer & AdamW \\
Hidden Dimension ($D$) & 1152 & Learning Rate & $1 \times 10^{-4}$ \\
Attention Heads & 16 & LR Scheduler & Cosine \\
MLP Expansion Ratio & 4.0 & Warmup Steps & 8,000 \\
Total Experts ($N$) & 4 & Total Fine-tuning Steps & 100,000 \\
Shared Experts & 1 & Global Batch Size & 64 \\
MoE Start Layer & 14 & EMA Decay Rate & 0.9999 \\
Auxiliary Loss Weight $\alpha$ & 0.01 & Precision & TF32 \\ \hline
\textbf{Diffusion Process} & \textbf{Value} & \textbf{Input / Output} & \textbf{Value} \\ \hline
Training Diffusion Steps & 1,000 & RGB Resolution & $256 \times 256$ \\
DDIM Sampling Steps & 16 & VAE Latent Dim & $32 \times 32 \times 4$ \\ \hline
\end{tabular}
\end{small}
\end{table}

\subsection{Denoising Diffusion Process}

\texttt{DeMUSE} utilizes a joint denoising objective $\mathbf{w}_0 = [\mathbf{z}_{t+1:t+H}, \mathbf{a}_{t:t+K-1}]$ to simultaneously model environmental evolution and action trajectories.

\begin{itemize}
    \item \textbf{Prediction Horizon and Synchronization}: We set the visual prediction horizon $H=3$ and the action sequence length $K=3$. Combined with a data sampling strategy of \texttt{skip\_step}=4, this configuration ensures that the generated \textit{visual imagination} remains temporally aligned with the $\sim$10 Hz control frequency of the robotic arm.
    
    \item \textbf{Noise Schedule and Sampling}: We adopt a linear noise schedule with $\beta$ values ranging from $1 \times 10^{-4}$ to $2 \times 10^{-2}$. During inference, deterministic DDIM sampling ($\eta=0$) with 16 steps is employed to achieve an optimal balance between generation fidelity and real-time execution efficiency.
    
    \item \textbf{Loss Weighting Strategy}: The model is optimized using Mean Squared Error (MSE) loss. To enhance the precision of robotic manipulation, we introduce an action loss weight $\lambda_{\text{action}}=2.0$.
\end{itemize}
    
\section{Datasets and Simulation Setup}
\label{appendix:data}

\subsection{Enhanced MetaWorld MT50 Platform}

To achieve physically consistent modeling for dexterous manipulation within a simulated environment, we implemented a \textbf{perceptual augmentation} layer for the MetaWorld platform. This framework integrates precise coordinate transformations with multi-modal signal preprocessing to ensure robust feature representation.

\begin{itemize}
    \item \textbf{End-Effector (EE) Frame Transformation}: By default, the MuJoCo physics engine outputs haptic signals in the global world frame $\mathcal{W}$. To ensure that sensory features are strictly aligned with the robot's action space and to enhance policy generalization, we decompose the 6-axis haptic signal $\mathbf{f}_t$ (as defined in Appendix \ref{appendix:arch}) into its 3-axis force component $\mathbf{f}_{\mathcal{W}}$ and 3-axis torque component $\boldsymbol{\tau}_{\mathcal{W}}$. We then perform a time-varying spatial transformation to the local EE-centric frame $\mathcal{E}$. Let the current EE orientation be represented by a rotation matrix $\mathbf{R}_{\mathcal{E}}^{\mathcal{W}} \in SO(3)$. The transformation logic is as follows:

\begin{enumerate}
    \item \textbf{Inverse Transformation}: Compute the world-to-local transformation matrix $\mathbf{R}_{\mathcal{W}}^{\mathcal{E}} = (\mathbf{R}_{\mathcal{E}}^{\mathcal{W}})^\top$, exploiting the orthogonality of rotation matrices.
    \item \textbf{Vector Space Projection}: Project the raw forces and torques onto the local EE manifold:
    \begin{equation}
    \mathbf{f}_{\mathcal{E}} = \mathbf{R}_{\mathcal{W}}^{\mathcal{E}} \cdot \mathbf{f}_{\mathcal{W}}, \quad \boldsymbol{\tau}_{\mathcal{E}} = \mathbf{R}_{\mathcal{W}}^{\mathcal{E}} \cdot \boldsymbol{\tau}_{\mathcal{W}}
    \end{equation}
    The transformed components are then re-concatenated to form the haptic token input for the Transformer backbone.
\end{enumerate}

    \item \textbf{Haptic Signal Preprocessing and Dynamic Range Emulation}: Contact forces in simulation often exhibit non-physical high-frequency artifacts due to solver constraints (e.g., penetration depth calculation). 
    \begin{enumerate}
        \item \textbf{Saturation Clipping and Outlier Rejection}: We implement bidirectional hard clipping (Force: $\pm 20$ N, Torque: $\pm 2$ Nm). This step emulates the finite dynamic range of industrial sensors while rejecting non-physical outliers that could disrupt the continuity of the latent feature distribution.
        \item \textbf{Savitzky-Golay (SG) Filtering}: Subsequent to clipping, a Savitzky-Golay filter with a window length of $w=5$ and polynomial order of 2 is applied. Compared to simple moving averages, the SG filter superiorly preserves the transient signal characteristics essential for detecting contact events.
    \end{enumerate}

    \item \textbf{Geometric Denoising of Depth Maps}: To mitigate "salt-and-pepper" noise—common in depth sensing due to occlusions, multiple reflections, or precision limits—we apply a $15 \times 15$ median filter to the $32 \times 32$ resolution depth maps. This operation effectively suppresses impulsive noise while preserving critical geometric edges, providing clean \textit{spatial structural anchors} for the subsequent Transformer blocks.
\end{itemize}

\subsection{Real-world Data Collection and Hardware Pipeline}
\label{appendix:realworld_pipeline}

To validate the generalization capabilities of DeMUSE in unstructured physical environments, we developed a high-fidelity embodied data collection system. The hardware suite comprises a UR10 collaborative robot, an OnRobot RG2-FT gripper integrated with a 6-axis Force/Torque (F/T) sensor, and an Intel RealSense D435i camera. To address the significant frequency discrepancies among heterogeneous sensors (camera at 30 Hz vs. F/T sensor at 100 Hz), we designed an advanced \textbf{asynchronous alignment pipeline}.

\begin{itemize}
    \renewcommand{\labelitemi}{$\bullet$}
    \item \textbf{Multi-threaded Asynchronous Acquisition}: The system employs an event-driven multi-threaded architecture to eliminate blocking latencies and ensure high-frequency responsiveness:
    \begin{enumerate}
        \item \textbf{Perception Thread}: The camera thread samples at 30 Hz, recording both hardware timestamps (for frame deduplication) and system monotonic timestamps (\texttt{cam\_ts\_mono}).
        \item \textbf{State Thread}: The robot thread operates at a 100 Hz loop, capturing the Tool Center Point (TCP) pose, gripper states, and raw 6-axis F/T signals.
        \item \textbf{Control \& Alignment Thread}: Driven at a 10 Hz frequency, this thread is responsible for retrieving and aligning multi-modal observation pairs from the buffers for inference.
    \end{enumerate}

    \item \textbf{Dual-Timestamp Alignment and Causal Consistency}: To ensure that the generated \textit{visual imagination} remains precisely synchronized with haptic feedback, we implemented a retrieval algorithm based on dual-timestamped circular buffers (ring buffers). All sensor data are mapped into a unified clock domain with a strict alignment threshold of $\Delta t = 50$ ms. Frames exceeding this threshold are invalidated, ensuring that the \textbf{observation-action sequence} ($o_t, a_t$) maintains a rigid causal chain undisturbed by system jitter.

    \item \textbf{Signal Normalization and Artifact Reduction}: Real-world haptic signals are acquired via the Modbus TCP protocol, with the mean value from the two sensors (left and right) used as the baseline. We apply a dynamic range clipping of $\pm 100$ N for force and $\pm 10$ Nm for torque to filter out outlier artifacts caused by mechanical vibrations. Subsequently, all signals undergo global \textbf{Z-score normalization} to ensure the feature distribution remains consistent with the simulation-based pre-training phase.
\end{itemize}

\subsection{Real-world Task Rollout Visualizations}
\label{appendix:realworld_results}

Building upon the core task definitions provided in the main text, this section provides sequential rollout visualizations to demonstrate the  performance of DeMUSE. 
\begin{figure*}[t]
  \centering
  \includegraphics[width=0.95\textwidth]{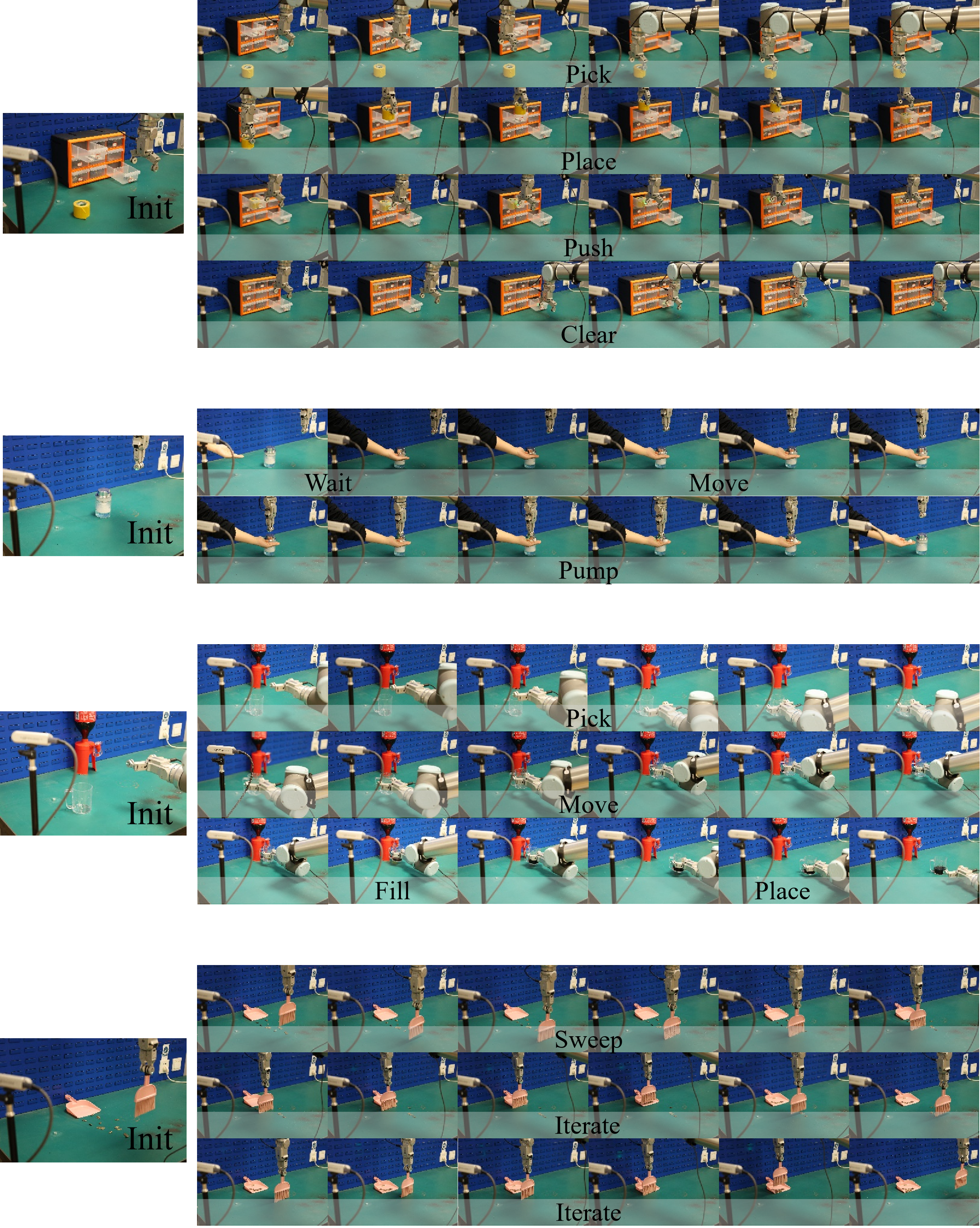} 
  \caption{\textbf{Qualitative execution sequences across real-world dexterous tasks.} The rows from top to bottom correspond to: \textsc{Tool Drawer Org.}, \textsc{Sweeping to Dustpan}, \textsc{Dispense Sanitizer}, and \textsc{Fill Cup (1/3 Cola)}.  }
  \vspace{-0.2in}
  \label{fig:realworld_visualization}
\end{figure*}

\end{document}